\documentclass[10pt, a4paper]{article}

\usepackage{lrec-coling2024}

\usepackage{inconsolata}
\usepackage{amsfonts,amssymb,amsmath}
\usepackage{multirow}
\usepackage{booktabs}
\usepackage{threeparttable}
\usepackage{algorithm}
\usepackage{algorithmicx}
\usepackage{algpseudocode}
\usepackage{graphicx}
\usepackage{subfigure}

\title{Fast Adversarial Training against Textual Adversarial Attacks}

\name{Yichen Yang$^{\ast}$\thanks{$^{\ast}$\quad The first two authors contribute equally.}, Xin Liu$^{\ast}$, Kun He$^{\dagger}$\thanks{ $^{\dagger}$\quad Corresponding author.}} 

\address{Huazhong University of Science and Technology, Wuhan, China\\
\{yangyc, liuxin\_jhl, brooklet60\}@hust.edu.cn\\}

\abstract{
Many adversarial defense methods have been proposed to enhance the adversarial robustness of natural language processing models. However, most of them introduce additional pre-set linguistic knowledge and assume that the synonym candidates used by attackers are accessible, which is an ideal assumption. We delve into adversarial training in the embedding space and propose a Fast Adversarial Training (FAT) method to improve the model robustness in the synonym-unaware scenario from the perspective of single-step perturbation generation and perturbation initialization. Based on the observation that the adversarial perturbations crafted by single-step and multi-step gradient ascent are similar, FAT uses single-step gradient ascent to craft adversarial examples in the embedding space to expedite the training process. Based on the observation that the perturbations generated on the identical training sample in successive epochs are similar, FAT fully utilizes historical information when initializing the perturbation. 
Extensive experiments demonstrate that FAT significantly boosts the robustness of BERT models in the synonym-unaware scenario, and outperforms the defense baselines under various attacks with character-level and word-level modifications.
 \\ \newline }

\begin{document}

\maketitleabstract

\section{Introduction}

Deep neural networks have been demonstrated to be vulnerable to adversarial examples~\cite{szegedy14intriguing,goodfellow14explaining,papernot16crafting}, which are crafted by adding imperceptible perturbation to the benign input. 
For Natural Language Processing (NLP) models, there are three types of adversarial attacks to generate adversarial examples with different perturbation granularity: character-level~\cite{gao18blackbox,hotflip18ebrahimi}, word-level~\cite{ren19generating,alzantot18generating,zang20word,li20bert}, and sentence-level attacks~\cite{wang19t3}. The word-level attacks based on synonym substitutions are most commonly used, as they basically guarantee the correct syntax, preserve unchanged semantics, and have a high attack success rate.  From another perspective 
of model visibility, adversarial attacks fall into two categories: white-box attacks and black-box attacks. White-box attacks~\cite{papernot16crafting, guo21gradient} have access to the model parameters, embeddings and gradients, while 
black-box attacks~\cite{jin20is,li20bert} only obtain the model output to generate adversarial examples and are more practical.

Many defense methods have been proposed to enhance the model's robustness against adversarial attacks based on synonym substitutions. However, we observe that most of them are synonym-aware, \emph{i.e.}, they assume all or part of the synonym substitutions used by the attackers are known beforehand when training the model, which is an ideal assumption. Firstly, as exhibited in the experiments of \citet{li2021searching} and \citet{wang2023robustness}, when the synonyms used by the defender are inconsistent with that of the attacker, the model's robustness will decline significantly. Secondly, attackers have various approaches to obtain synonym candidates, such as artificially formulating the embedding distance~\cite{alzantot18generating,jin20is}, retrieving synonyms through the online thesaurus~\cite{ren19generating,zang20word}, or inferring by language models~\cite{li20bert}. It is difficult to define all possible synonym substitutions 
for textual adversarial defenses. Therefore, in this paper, we focus on a more realistic scenario where the defense method is inaccessible to synonym information or any additional human-prescribed linguistic knowledge 
other than the dataset. 

We rethink the Adversarial Training (AT) methods in the synonym-unaware scenario. As a typical kind of defense method to improve the model's robustness, most AT methods, such as ATFL~\cite{wang21adversarial} and BFF~\cite{ivgi2021achieving}, work in the input space and utilize a specific adversarial attack to generate adversarial texts to train the model. 
They require pre-introduced synonym information to craft the adversarial texts. ASCC~\citep{dong21towards} trains models with the virtual adversarial examples constructed by the combination of the embedding representation of synonyms, and it still needs the synonym information. In contrast, another type of AT method~\cite{adversarial17miyato,zhu20freelb,adversarial20liu} works in the embedding space and have no need to access to synonyms. They perturb the embedding representation directly and train the model using the perturbed embedding representation. However, their goal is primarily to improve the model generalization on the original test set as a regularization technique rather than boosting the adversarial robustness. Experiments conducted by \citet{liu22flooding} and \citet{li21token} also demonstrate that such type of AT method has a weak effect on improving the model's robustness.

In this work, we empirically demonstrate that AT in the embedding space could also improve the model's robustness without additional synonym information or human-prescribed linguistic knowledge. Generally, the Projected Gradient Decent (PGD) method~\cite{madry18towards} is adopted to generate adversarial perturbation on the embedding representation. However, since the PGD attack is based on multi-step gradient ascent, PGD-AT is 
very inefficient for commonly used large-scale pre-trained NLP models such as BERT~\cite{devlin19bert}, leading to unsatisfactory performance within a limited time. To address this issue, we propose a Fast Adversarial Training (FAT) method to boost the model's robustness from the perspective of single-step perturbation generation and  
initialization with historical information.

Firstly, we observe that the adversarial perturbations crafted by single-step and multi-step gradient ascent are similar for NLP models. Therefore, FAT adopts single-step gradient ascent to craft perturbation on the embedding representation instead of using multi-step gradient ascent to expedite the training process. Then, the model could be trained with more epochs to achieve better robustness within a limited time.

Secondly, we observe that the direction of the perturbations generated on the identical training sample in two successive training epochs is similar. To make full use of the historical information, FAT initializes the perturbation along the direction of perturbation generated on the same sample in the previous epoch.


Besides, we argue that it is unnecessary to generate perturbation for each training sample in each epoch. We propose a variant method, termed FAT-\(I\), in which the adversarial perturbation of each training sample is only updated once in multiple epochs, further accelerating the training process. 

Extensive experiments on three benchmark datasets show that our proposed FAT achieves the best robustness under various advanced adversarial attacks involving different model visibility and perturbation granularity.

Our main contributions are as follows:
\begin{itemize}

\item By observing the characteristics of perturbation generation in the embedding space of NLP models, we propose Fast Adversarial Training (FAT), which adopts single-step gradient ascent to expedite the training process and fully 
utilizes historical training information to achieve better robustness.

\item We further propose the variant AT method FAT-\(I\), which accelerates the training process without much decay in the model's robustness.

\item Since textual adversarial attacks have various settings of synonym candidates and perturbation budgets, it is valuable for FAT to provide an easy-to-apply and effective solution for adversarial defense under the realistic synonym-unaware scenario. Furthermore, Extensive experiments demonstrate that FAT and its variants achieve the best robustness among the defense baselines with a clear margin.

\end{itemize}

\section{Related Work}

Many adversarial defense methods have been proposed to boost the model's robustness against adversarial attacks based on synonym substitutions. Here we divide these methods into two categories, \emph{i.e.}, synonym-aware methods and synonym-unaware methods.

Most defense methods need to be accessible to the synonyms used by attackers or introduce human-prescribed rules to determine synonyms. Input transformation methods encode the synonyms to the same code~\cite{wang21natural} or adopt synonym substitutions~\cite{mozes21frequency} to eliminate the perturbation in the input space. \citet{yang22robust} embrace
the triplet metric learning to pull words closer to their synonyms and push away their non-synonyms in the embedding space. 
Interval bound propagation methods~\cite{jia19certified,wang2023robustness} compute the interval of all possible perturbed texts under a particular synonym definition and propagate the interval bounds through the network layers to minimize loss in the worst case. Some certified methods~\cite{zhao22certified,ye20safer} utilize randomized smoothing to achieve provable robustness. Adversarial training (AT) methods working in the input space~\cite{wang21adversarial,ivgi2021achieving} craft adversarial texts based on synonym substitutions and regard adversarial texts as the training data. It is worth noting that although some AT methods involve the embedding space to generate adversarial perturbation, our work differs from those AT methods. Specifically, ATFL~\citep{wang21adversarial} uses the gradient \textit{w.r.t.} embedding as guidance to find the important words to perturb, and it trains models with adversarial texts rather than perturbed embedding representation. ASCC~\citep{dong21towards} trains models with the virtual adversarial examples constructed by the combination of the embedding representation of synonyms. The above two AT methods still rely on the synonyms and differ from our synonym-unaware approach.

Since the synonym-unaware defense methods  do not involve synonym information used by the attackers, it is more consistent with the realistic scenario, and we can evaluate their robustness more fairly. Flooding-X~\cite{liu22flooding} leverages the Flooding method to improve the model's robustness, 
that is a simple training strategy to avoid zero training loss and guide the model into a smooth parameter landscape. InfoBERT~\cite{wang21infobert} introduces two mutual information based regularizers for model training. A series of works~\cite{adversarial17miyato,zhu20freelb,li21token} directly perturb the word embeddings and utilize the perturbed embedding representation to train the model. However, these works regard AT as a regularization strategy and aim to improve the model's 
generalization on the original dataset rather than adversarial robustness.
\citet{li2021searching} utilize PGD without projection operation to add a large magnitude of perturbation to the embedding representations for AT, which is the most similar work to ours. In contrast, we utilize single-step gradient ascent to generate adversarial perturbation. Besides, \citet{li2021searching} randomly initialize the adversarial perturbation, while we introduce historical information in perturbation initialization for achieving better model's robustness. 

\section{Methodology}
In this section, we investigate adversarial training in the embedding space and present two observations. Based on the two observations, we propose a Fast Adversarial Training (FAT) method to boost the model's robustness.
We also describe the variants of FAT to further accelerate the training process.

\subsection{Rethinking Adversarial Training}

According to the position of perturbation, AT for NLP models could be divided into two categories, \emph{i.e.}, discrete AT~\cite{ivgi2021achieving,wang21adversarial} and continuous AT~\cite{li2021searching,li21token}. The former generates adversarial texts in the discrete input space, while the latter adds adversarial perturbations to the embedding representation in the continuous embedding space.

Given a dataset \(\mathcal{D}\) and a classification model \(f_{\boldsymbol{\theta}}(\cdot)\) parameterized by \(\boldsymbol{\theta}\), 
the training objective of the discrete AT could be formulated as:
\begin{equation}
    \min_{\boldsymbol{\theta}} \sum_{(x,y)\in \mathcal{D}}\mathcal{L}(f_{\boldsymbol{\theta}}(a(x)), y),
\end{equation}
where \(x\) denotes an input text with true label \(y\), \(a(x)\) is the adversarial text generated by a certain attack method \(a(\cdot)\), and \(\mathcal{L}(\cdot, \cdot)\) is the cross-entropy loss. The pre-set human-prescribed linguistic knowledge about synonyms needs to be introduced when generating adversarial texts. When the synonym candidates used in AT and the adversarial attack for evaluation are inconsistent, the performance of AT will decline significantly~\cite{li2021searching}.

In contrast, continuous AT needs no pre-set linguistic knowledge, and the training objective is:
\begin{equation}
\label{eq:con_at}
    \min_{\boldsymbol{\theta}} \sum_{(x,y)\in \mathcal{D}}\left[\max_{\left \Vert \boldsymbol{\delta} \right \Vert_p \leq \epsilon} \mathcal{L}(f_{\boldsymbol{\theta}}(\boldsymbol{v}(x)+\boldsymbol{\delta}),y)\right],
\end{equation}
where \(\boldsymbol{v}(x)\) denotes the embedding representation of text \(x\), and \(\boldsymbol{\delta}\) is the adversarial perturbation added to the embedding representation. \(\left \Vert \cdot \right \Vert_p\) denotes \(l_p\)-norm, and \(\epsilon\) controls the perturbation magnitude. \citet{li2021searching} adopt \(l_2\)-norm PGD to solve the inner maximization in Eq.~(\ref{eq:con_at}) for training, which we call PGD-AT for simplicity. For the specific implementation, they remove the projection operation and iteratively conduct multiple steps of gradient ascent to generate the adversarial perturbation:
\begin{equation}
\label{eq:delta_init}
    \boldsymbol{\delta}^0 = \boldsymbol{U}(-\epsilon_0, \epsilon_0) / \sqrt{n \cdot d},
\end{equation}
\begin{equation}
    \boldsymbol{\delta}^{t+1} = \boldsymbol{\delta}^t + \alpha \cdot \frac{\nabla_{\boldsymbol{\delta}^t} \mathcal{L}(f(\boldsymbol{v}(x)+\boldsymbol{\delta}^t), y)}{\left \Vert \nabla_{\boldsymbol{\delta}^t} \mathcal{L}(f(\boldsymbol{v}(x)+\boldsymbol{\delta}^t), y) \right \Vert_2},
\end{equation}
where \(n\) is the number of words in the input text \(x\) and \(d\) is the dimension of word embedding. \(\boldsymbol{U}(-\epsilon_0, \epsilon_0) \in \mathbb{R}^{n \times d}\) denotes a matrix whose elements are uniformly sampled from \(-\epsilon_0\) to \(\epsilon_0\). \(t\) denotes the current step. \(\epsilon_0\) and \(\alpha\) are hyper-parameters for controlling the magnitude of initial perturbation and  step size, respectively.

However, since the current commonly used NLP models are large-scale pre-training models, such as BERT~\cite{devlin19bert}, using PGD attack to generate adversarial examples for AT is 
very inefficient.
For instance, on the binary classification dataset \textit{IMDB} which contains 25,000 training samples, PGD-AT takes about two hours to train a BERT model of the base version for one epoch on a single TITAN RTX GPU.
Worse still, adversarial examples are more diverse than benign samples. 
Thus, PGD-AT is hard to converge to satisfactory robustness within a limited time.

\begin{figure}
    \centering
    {
    \includegraphics[width=\columnwidth]{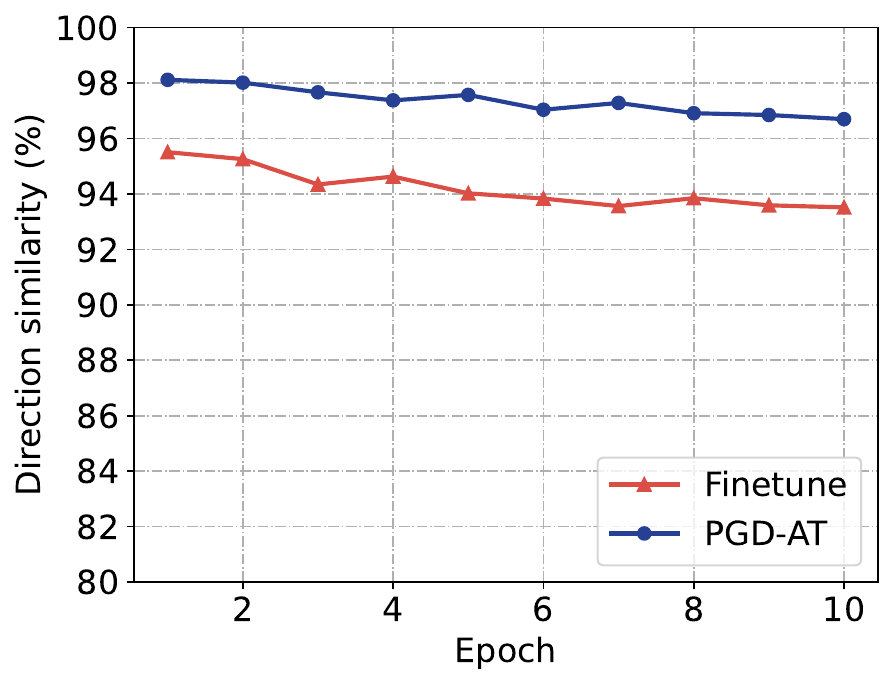}}
    \vspace{-1em}
    \caption{The direction similarity between perturbations generated by single-step and multi-step gradient ascent in the embedding space.}
    \label{fig:comp_fgsm_pgd}
\end{figure}

\subsection{Fast Adversarial Training}
\label{sec:fat}

We propose a Fast Adversarial Training (FAT) method to enhance the defense performance of continuous AT from the perspective of single-step perturbation generation and 
initialization with historical information.

\subsubsection{Single-Step Perturbation Generation} 

We speculate that it is 
redundant to adopt multi-step gradient ascent to generate adversarial perturbation for AT on NLP models. 
For validation, we randomly choose 1000 testing samples from the \textit{IMDB} dataset and use single-step gradient ascent and multi-step gradient ascent, respectively, to generate adversarial perturbations on standard fine-tuning and PGD-AT trained models. We apply the element-wise sign function to the two perturbations as their directions. The direction similarity (\%) could be defined as the ratio of the number of dimensions with the same value between the two directions to the total number of dimensions. 
As illustrated in Figure~\ref{fig:comp_fgsm_pgd}, on models trained by standard fine-tuning or PGD-AT, we observe that the direction similarity between the perturbations generated by 
multi-step and single-step gradient ascents 
is the same over 90\% of the dimensions,
especially on the PGD-AT model. This phenomenon indicates that multi-step perturbation generation is 
redundant for AT of NLP models.

We thereby adopt the single-step gradient ascent to generate adversarial perturbation to boost the efficiency of AT. Specifically, with the initial adversarial perturbation \(\boldsymbol{\delta}^0\), the training objective could be formulated as follows:
\begin{equation}
\label{eq:fat}
    \boldsymbol{\delta} = \boldsymbol{\delta}^0 + \epsilon \cdot \frac{\nabla_{\boldsymbol{\delta}^0} \mathcal{L}(f(\boldsymbol{v}(x)+\boldsymbol{\delta}^0), y)}{\left \Vert \nabla_{\boldsymbol{\delta}^0} \mathcal{L}(f(\boldsymbol{v}(x)+\boldsymbol{\delta}^0), y) \right \Vert_2},
\end{equation}
\begin{equation}
    \min_{\boldsymbol{\theta}} \sum_{(x,y)\in \mathcal{D}} \mathcal{L}(f_{\boldsymbol{\theta}}(\boldsymbol{v}(x)+\boldsymbol{\delta}),y).
\end{equation}

In the AT process of large-scale NLP models such as BERT, most of the time 
cost is due to the gradient back-propagation. Assuming that the number of training samples is \(N\), the number of training epochs is \(E\), and the step of adversary generation is \(T\), PGD-AT requires \(NE(T+1)\) back-propagation, in which each training sample needs \(T\) back-propagation to generate adversarial perturbation in each epoch, and one back propagation to update model parameters. In contrast, FAT requires only \(2NE\) back-propagation. Therefore, within a given limited time, FAT can conduct more training epochs and achieve higher robustness.

\begin{figure}
    \centering
    {
    \includegraphics[width=\columnwidth]{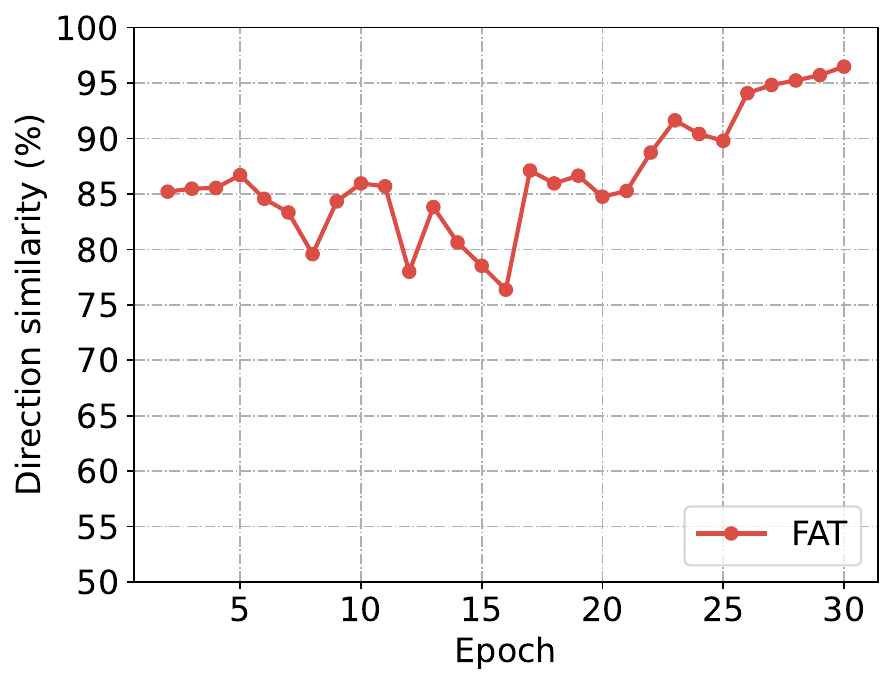}}
    \vspace{-1em}
    \caption{The direction similarity between perturbations generated by last and current epoch in the embedding space.}
    \label{fig:comp_fgsm_epoch}
\end{figure}

\subsubsection{Perturbation Initialization}

As in Eq.~(\ref{eq:delta_init}), previous AT methods~\cite{li21token,li2021searching} introduce randomness into training data by initializing the perturbation with a small random noise. 
We argue that introducing useful information for initialization could help craft the adversarial examples from a good point for training and enhance the model's robustness.

For validation, we randomly choose 1000 testing samples from the \textit{IMDB} dataset and use single-step gradient ascent to generate adversarial perturbations for each epoch of FAT. As illustrated in Figure~\ref{fig:comp_fgsm_epoch}, we observe that in the training course of FAT, the direction of adversarial perturbation generated on the same training sample in two successive epochs is identical in 77\%-97\% of the dimensions, indicating that the adversarial perturbation generated in the previous epoch contains helpful information for the next epoch.

To fully use historical perturbation, we propose a new initialization approach to introduce randomness into training data and find a better starting point for generating the adversarial examples. 
Specifically, we limit the initial perturbation \(\boldsymbol{\delta}_0\) to the perturbation direction corresponding to the identical sample in the previous epoch, and the magnitude on each dimension is generated randomly, which could be formulated as follows:
\begin{equation}
\label{eq:our_init}
    \boldsymbol{\delta}^0 = \boldsymbol{U}(0, \epsilon_0) \odot \mathrm{sign} (\boldsymbol{\delta}') / \sqrt{n\cdot d},
\end{equation}
where \(\odot\) denotes element-wise multiplication, and \(\mathrm{sign}(\cdot)\) denotes the element-wise sign function. \(\boldsymbol{\delta}'\) is the perturbation of the identical training sample in the previous epoch.

In this way, we could incorporate information of historical perturbation into the current epoch, similar to the idea of momentum, which can stabilize the generation of adversarial examples and improve the model's robustness. 

\begin{algorithm}
    \caption{FAT-\(I\)}
    \label{alg:FAT}
    \begin{algorithmic}
    \State {\bfseries Input:} Training data \(\mathcal{D}\), model \(f_{\boldsymbol{\theta}}\), initial perturbation size \(\epsilon_0\), perturbation size \(\epsilon\), perturbation update interval \(I\), number of training epochs \(E\)
    \State {\bfseries Output:} Robust model \(f_{\boldsymbol{\theta}}\)
    \For{\(i=1,2,\cdots,\left | \mathcal{D}\right|\)}
    \State \(\boldsymbol{\delta}_i \leftarrow \boldsymbol{U}(-\epsilon_0, \epsilon_0) / \sqrt{n \cdot d}\)
    \EndFor
    \For{$e=0,1, \cdots, E-1$}
    \For{\(\{(x_i, y_i)\} \subset \mathcal{D}\)}
    \If{\(e\%I=0\)}
    \State \textit{\# Update adversarial perturbations}
    \State {\(\boldsymbol{\delta}_i^0  \leftarrow  \boldsymbol{U}(0, \epsilon_0) \odot \mathrm{sign} (\boldsymbol{\delta}_i) / \sqrt{n\cdot d}\)}
    \State {\(\boldsymbol{\delta}_i \leftarrow \boldsymbol{\delta}_i^0 + \epsilon \cdot \frac{\nabla_{\boldsymbol{\delta}_i^0} \mathcal{L}(f(\boldsymbol{v}(x_i)+\boldsymbol{\delta}_i^0), y_i)}{\left \Vert \nabla_{\boldsymbol{\delta}_i^0} \mathcal{L}(f(\boldsymbol{v}(x_i)+\boldsymbol{\delta}_i^0), y_i) \right \Vert_2}\)}
    \EndIf
    \State {Compute loss \(\mathcal{L}(f_{\boldsymbol{\theta}}(\boldsymbol{v}(x_i)+\boldsymbol{\delta}_i), y_i)\) 
    \State Update model parameters \(\boldsymbol{\theta}\)}
    \EndFor
    \EndFor
    \State \textbf{return} $f_{\boldsymbol{\theta}}$
    \end{algorithmic}
\end{algorithm}

\subsection{Variants of FAT}
As illustrated in Figure~\ref{fig:comp_fgsm_epoch}, the adversarial perturbations generated on the identical training sample in two successive epochs are similar. This phenomenon also indicates that instead of generating adversarial perturbation for each training sample at each epoch, we can craft adversarial examples using the perturbations of the previous epoch, further accelerating the training process.

Therefore, we propose the variant Fast Adversarial Training with Interval (FAT-\(I\)), which is summarized in Algorithm~\ref{alg:FAT}. Specifically, we update and record the adversarial perturbation of each training sample in the \(Ik\)-th epoch, where \(I>0\) is the update interval and \(k=0,1,2\cdots\). In other epochs, we directly add the recorded perturbation to the embedding representation of the corresponding training sample. When \(I=1\), it is equivalent to FAT. 

\section{Experiments}
\begin{table}
\centering
\resizebox{\columnwidth}{!}{
\setlength{\tabcolsep}{1mm}{
\begin{threeparttable}
\begin{tabular}{lcccc}
\toprule
Dataset & \#Training & \#Testing & \#Class & \ Avg. words\\
\midrule
\textit{IMDB} & ~25k & ~25k & 2 & ~~~~268 \\
\textit{AGNEWS} & 120k & 7.6k & 4 & ~~~~~~40 \\
\textit{QNLI} & 105k & 5.4k & 2 & 11 / 31\tnote{*} \\
\bottomrule
\end{tabular}
\end{threeparttable}
}}

\caption{Statistics of datasets. $^*$ denotes the average words of premise and hypothesis.
}
\label{tab:datasets}
\end{table}
This section evaluates the robustness of the proposed FAT and four typical defense baselines against various adversarial attacks on BERT model across three datasets. Then, we investigate the performance of the variant FAT-\(I\) with different values of \(I\). Besides, we provide hyper-parameter analysis on the perturbation magnitude and training epochs.
\begin{table*}[t]
    \centering
    \begin{threeparttable}
    \begin{tabular}{c|l|c|cc|cc|cc}
        \toprule
         \multirow{2}{*}{Dataset} & \multirow{2}{*}{Defense} & \multirow{2}{*}{\textit{Clean\%}}  & \multicolumn{2}{c}{TextFooler} & \multicolumn{2}{c}{BERT-Attack} & \multicolumn{2}{c}{TextBugger}\\
         \cmidrule(lr){4-5} \cmidrule(lr){6-7} \cmidrule(lr){8-9}
          ~ & ~ & ~ & \textit{Aua\%} & \textit{\#Query} & \textit{Aua\%} & \textit{\#Query} & \textit{Aua\%} & \textit{\#Query} \\
         \midrule
         \multirow{7}{*}{\textit{IMDB}} &
         Finetune\tnote{*}   & 95.0 & 24.5  & 1533.15  &  20.3  &  2237.38  &  48.7  &   1160.35 \\
         &PGD-AT\tnote{*}   & 95.0 & 26.3   & 1194.08  & 21.3  & 1465.83  &  52.3  &   ~~982.02  \\
         &TAVAT\tnote{*}  & 95.5 & 27.6 & 1205.80 & 23.1 & 2244.77 & 54.1 & 1022.56 \\
         &InfoBERT \tnote{*}  & 96.3 & 27.4 & 1094.55 & 20.8 & 1428.67 & 49.8 & 1215.39 \\
         & Flooding-X\tnote{*}  & \textbf{97.5} & 40.5   & 2315.35   & 32.3   &  2248.71  & 62.3   & \textbf{2987.95}  \\
         & FAT (w/o)   & 94.9  &  67.3  & 2550.85   &  49.8  &  3503.30  &  70.4  & 1650.51  \\
         & FAT   & 95.0  &  \textbf{70.8}  & \textbf{2574.45}   &  \textbf{55.1}  &  \textbf{3636.75}  &  \textbf{75.0}  & 1687.15  \\
         \midrule
         \multirow{7}{*}{\textit{AGNEWS}} &
         Finetune\tnote{*}    & 94.9  & 20.5 & ~~372.14  & ~~6.5  & ~~477.34  & 42.7   & ~~192.75  \\
         &PGD-AT\tnote{*}   & 94.8  &  37.2   & ~~428.13  & 32.8  & ~~704.78  &  58.2  & ~~252.87   \\
         &TAVAT\tnote{*} & \textbf{95.2} &39.7&~~441.11&23.7&~~672.52&55.9&~~234.01 \\
         &InfoBERT\tnote{*} &94.6&29.2&~~406.32&15.6&~~598.25&50.7&~~201.66\\
         & Flooding-X\tnote{*}   & 94.9  &  42.4    & ~~451.35  &  27.4 & ~~690.27  &  62.2  &  ~~222.49 \\
         & FAT (w/o)  & \textbf{95.2} &  60.8   & ~~500.62  & \textbf{48.6}  & \textbf{~~764.88}  &  \textbf{65.9}  & \textbf{~~306.94}  \\
         & FAT  & 95.1 & \textbf{62.3} & \textbf{~~505.86}  &   48.0  &  ~~754.63  & 63.6   & ~~301.91  \\
         \midrule
         \multirow{7}{*}{\textit{QNLI}} &
         Finetune\tnote{*} & 90.6  & ~~5.3   & ~~161.88  & ~~3.5  & ~~216.46  & 10.9   & ~~~~98.39  \\
         &PGD-AT\tnote{*}  & 90.6  & 28.1   & ~~269.38  & 24.0  &  ~~399.91 &  33.8  &  ~~154.55  \\
         & InfoBERT\tnote{*} & 90.4 & 23.1 & ~~250.87 & 11.1 & ~~268.91 & 12.8 & ~~127.93\\
         & Flooding-X\tnote{*}  & \textbf{91.8}  &  27.9  & ~~251.17  &  26.2 & ~~364.06  &  29.5  &  ~~137.12 \\
         & FAT (w/o) & 91.4 & 44.8 &~~271.69  &  28.1  & ~~384.83  & 39.4  &  ~~172.35    \\
         & FAT & 91.1 & \textbf{48.3} &  \textbf{~~280.07}  &  \textbf{33.0}  & \textbf{~~414.37}   &  \textbf{44.3}  & \textbf{~~184.29}   \\
         \bottomrule
    \end{tabular}
    \end{threeparttable}
    \caption{The comparison results of FAT and baselines under various adversarial attacks on BERT model. FAT (w/o) denotes FAT using random perturbation initialization rather than our proposed initialization method. $^*$ indicates results reported in \citet{liu22flooding}.
    The best performance is highlighted in \textbf{bold}.}
    \label{tab:main_results}
\end{table*}
\begin{table}[t]
\centering
\begin{tabular}{c|l|c}
\toprule
Dataset & Model & \textit{Aua (\%)}\\ 
\midrule
\multirow{3}{*}{\textit{IMDB}} & Finetune & ~~0.4 \\
~ & Flooding-X & 40.0  \\
~ & FAT & \textbf{46.8}  \\
\midrule
\multirow{3}{*}{\textit{AGNEWS}} & Finetune & ~~0.4\\
~ & Flooding-X &16.3   \\
~ & FAT &\textbf{26.5} \\
\midrule
\multirow{3}{*}{\textit{QNLI}} & Finetune &16.3 \\
~ & Flooding-X &\textbf{47.8}  \\
~ & FAT &42.3\\
\bottomrule
\end{tabular}
\caption{The defense performance of FAT and baselines under the GBDA attack, which is a white-box attack and has no metric for the number of queries.}
\label{tab:more_attack}
\end{table}

\subsection{Experimental Setup}
\subsubsection{Tasks and Models} 

To thoroughly evaluate the effectiveness of the proposed method, we conduct experiments on the text classification task of \textit{IMDB}~\cite{maas11learning} and \textit{AGNEWS}~\cite{zhang15character} datasets, and the natural language inference task of \textit{QNLI}~\cite{wang19glue} dataset. The three standard benchmark datasets have various text lengths, number of classes, and sample sizes. Their specific information is shown in Table~\ref{tab:datasets}.
We train the BERT model~\cite{devlin19bert} of the uncased base version on the three datasets. 

\subsubsection{Attack Methods} 

Since we focus on the defense without any additional pre-set linguistic knowledge and synonym information, we utilize four adversarial attacks, namely TextFooler~\cite{jin20is}, BERT-Attack~\cite{li20bert}, TextBugger~\cite{li19textbugger}, GBDA~\cite{guo21gradient}, involving character-level perturbations and word-level perturbations based on different synonym candidates. TextFooler defines synonyms based on the cosine distance between the word vectors, then identifies important words in the input text and performs synonym substitutions on them. BERT-Attack utilizes pre-trained masked language models to mine for synonym candidates. TextBugger mixes the character-level and word-level perturbations to attack the model. GBDA is a challenging white-box attack, which searches for a distribution of adversarial examples parameterized by a continuous valued matrix and unitizes gradient-based
optimization to craft adversarial examples.
For the first three attacks, we use the default implementation in \textsc{TextAttack}\footnote{https://github.com/QData/TextAttack}. For GBDA attack, we use the implementation of the paper\footnote{https://github.com/facebookresearch/text-adversarial-attack}. For the natural language inference task, each sample consists of two sentences: the premise and the hypothesis, and typically, only the hypothesis is perturbed to keep the true label unchanged. We randomly choose 800 test samples from each dataset to generate adversarial examples.

\subsubsection{Defense Baselines}

Following~\citet{liu22flooding}, we compare our method with standard finetuning~\cite{devlin19bert} and four defense baselines, PGD-AT~\cite{madry18towards,li2021searching}, TAVAT~\cite{li21token}, InfoBERT~\cite{wang21infobert}, and Flooding-X~\cite{liu22flooding}. All the baselines and our proposed method require no additional pre-set linguistic knowledge and can therefore be evaluated fairly.

\subsubsection{Training Details}

Our implementations are based on~\citet{liu22flooding}
in which PGD-AT is run with 5 attack steps for 10 epochs. According to the analysis in Section~\ref{sec:fat}, we run our proposed FAT for 30 epochs to achieve the same time consumption as PGD-AT, among which the last epoch is selected for the evaluation. For hyper-parameters in Eq.~(\ref{eq:our_init}) and Eq.~(\ref{eq:fat}), we set \(\epsilon_0=\) 0.05 and \(\epsilon=\) 0.2.



\begin{table*}[t]
    \centering
    \begin{threeparttable}
    \begin{tabular}{c|l|c|c|cc|cc|cc}
        \toprule
         \multirow{2}{*}{Dataset} & 
         \multirow{2}{*}{Defense} & 
        \multirow{2}{*}{\textit{Time}} & 
         \multirow{2}{*}{\textit{Clean\%}} & 
         \multicolumn{2}{c}{TextFooler} & \multicolumn{2}{c}{BERT-Attack} & \multicolumn{2}{c}{TextBugger}\\
         \cmidrule(lr){5-6} \cmidrule(lr){7-8} \cmidrule(lr){9-10}
          ~ & ~ & ~ & ~ & \textit{Aua\%} & \textit{\#Query} & \textit{Aua\%} & \textit{\#Query} & \textit{Aua\%} & \textit{\#Query} \\
         \midrule
         \multirow{3}{*}{\textit{IMDB}} &
         FAT & 2/2 &\textbf{95.0}  &  \textbf{70.8}  & \textbf{2574.45}   &  \textbf{55.1}  &  \textbf{3636.75}  &  \textbf{75.0}  & \textbf{1687.15}  \\
         & FAT-2 & 3/4 & \textbf{95.0}  &  68.8  &   2367.49  & 50.0   & 3523.00  & 70.1 & 1536.36\\
         & FAT-3 & \textbf{4/6}  &94.8  & 64.0   &  2322.80  & 43.1   & 3097.64   & 68.1   & 1514.06\\
         \midrule
         \multirow{3}{*}{\textit{AGNEWS}} &
         FAT  & 2/2 & 95.1 & \textbf{62.3} & \textbf{~~505.86} & 48.0 & \textbf{~~754.63} & 63.6 & ~~301.91 \\
         & FAT-2 &3/4& \textbf{95.3} & 60.4 & ~~492.11 & \textbf{48.5} & ~~750.22 & \textbf{64.4} & ~~300.93 \\
         & FAT-3 &\textbf{4/6} & 95.2 & 60.1 & ~~490.21 & 47.8 & ~~745.34 & 63.3 & \textbf{~~302.06}\\
         \midrule
         \multirow{3}{*}{\textit{QNLI}} &
         FAT &2/2 & 91.1 & 48.3 & ~~280.07 & 33.0 & \textbf{~~414.37} & \textbf{44.3} & \textbf{~~184.29}  \\
         & FAT-2 &3/4& \textbf{91.5} & \textbf{48.4} & \textbf{~~281.58} & \textbf{33.9} & ~~411.64 & 43.5 & ~~183.20 \\
         & FAT-3&\textbf{4/6} & 91.4 & 45.8 & ~~275.82& 32.5& ~~396.74& 40.9& ~~174.96\\
         \bottomrule
    \end{tabular}
    \end{threeparttable}
    \caption{The comparison results of the variant FAT-\(I\) with different values of \(I\) under various attacks on BERT model. \textit{Time} denotes the relative training time of FAT-\(I\) and FAT. The best 
    appears in \textbf{bold}.
    }
    \label{tab:main_interval}
\end{table*}
\subsection{Main Results}
\label{mainResults}
We compare FAT with standard finetuning and four defense baselines concerning the robustness against various attacks. 
The comparison results are shown in Table~\ref{tab:main_results}, which includes three evaluation metrics. 
\textit{Clean\%} denotes the classification accuracy on the entire original test set. \textit{Aua\%} is short for the accuracy under attacks. \textit{\#Query} denotes the average number of queries to attack each sample. The more effective the defense method, the higher the metrics of \textit{Aua\%} and \textit{\#Query}. 
Meanwhile, we also need to ensure that \textit{Clean\%} does not decline much compared to the standard finetuning. 


The results indicate that FAT has substantially enhanced the model's robustness, surpassing the defense baselines with a prominent margin on all the three datasets under various attacks, especially the TextFooler attack. For instance, FAT outperforms the best defense baseline by 30.3\%, 22.8\%, and 12.7\% on \textit{IMDB} dataset under the three attacks, respectively. Besides, FAT achieves the same or even improved accuracy on the original test set compared to standard finetuning.


Under the same time limit, FAT using single-step gradient ascent to generate perturbation performs better than PGD-AT using multi-step gradient ascent, which the following two reasons can probably explain. First, the difference between the perturbations generated by single-step and multi-step gradient ascent is trivial for AT on NLP models. Second, with much fewer calculations for each epoch, FAT run more epochs to achieve better robustness within a limited time.

In Table~\ref{tab:main_results}, FAT (w/o) uses random perturbation initialization rather than our method. The comparison between FAT and FAT (w/o) reveals that the initialization with previous information is crucial for crafting adversarial examples. For instance, when we initialize the perturbation along the direction of the previously generated one, FAT performs better than FAT (w/o) under the TextFooler attack on the three datasets, with the improvement of 3.5\%, 1.5\%, and 3.5\%, respectively.

\subsection{Evaluation with White-box Attack}

To further verify the effectiveness of defense, we compare FAT with 
standard finetuning and the best baseline method Flooding-X on the GBDA attack~\cite{guo21gradient}, which is a more challenging white-box attack.

The results are shown in Table~\ref{tab:more_attack}. The GBDA attack severely degrades the accuracy of finetuned BERT models. FAT have enhanced the model's robustness by a clear margin of 46.4\%, 26.1\%, and 26.0\% on the three datasets, respectively, indicating that not only in the black-box scenario but also in the white-box scenario, FAT can provide an essential defense against adversarial examples. 

\subsection{Variants of FAT}

The variant FAT-\(I\) updates the perturbation once for every \(I\) epochs to further expedite the training process. The training time required by FAT-\(I\) is about \((I+1)/2I\) of that of FAT, which is estimated by the number of back-propagation during the training. The larger \(I\) is, the less training time be required. Note that FAT-\(I\) is equivalent to FAT when \(I=1\). We investigate the performance when \(I=1,2,3
\), and the results are reported in Table~\ref{tab:main_interval}.


Generally, a larger \(I\) shortens the training time without much decay in model's robustness under various attacks. For instance, when \(I=2\), the training time of FAT-\(2\) is \(3/4\) of that of FAT, and the performance of FAT-\(2\) concerning robustness is comparable to FAT and even better than FAT in some cases. Furthermore, FAT-\(3\) still outperforms all the defense baselines in the three datasets (See results in Table~\ref{tab:main_results}). Besides, the value of \(I\) has little effect on the accuracy of the original test set, which is constant or slightly increases as \(I\) becomes larger.

To sum up, FAT-\(I\) provides a flexible option to achieve a more proper trade-off between training time and robustness.



\begin{figure*}[tb]
    \centering
    \subfigure[ \textit{IMDB}]{\label{fig:epsilon:imdb}\includegraphics[width=.32\textwidth]{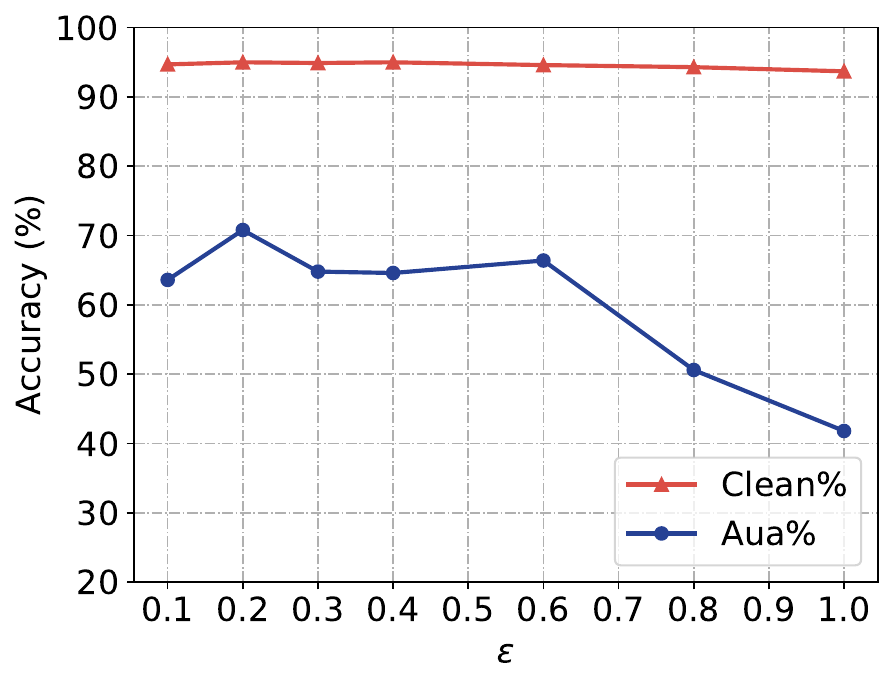}}
    \subfigure[ \textit{AGNEWS}]{\label{fig:epsilon:agnews}\includegraphics[width=.32\textwidth]{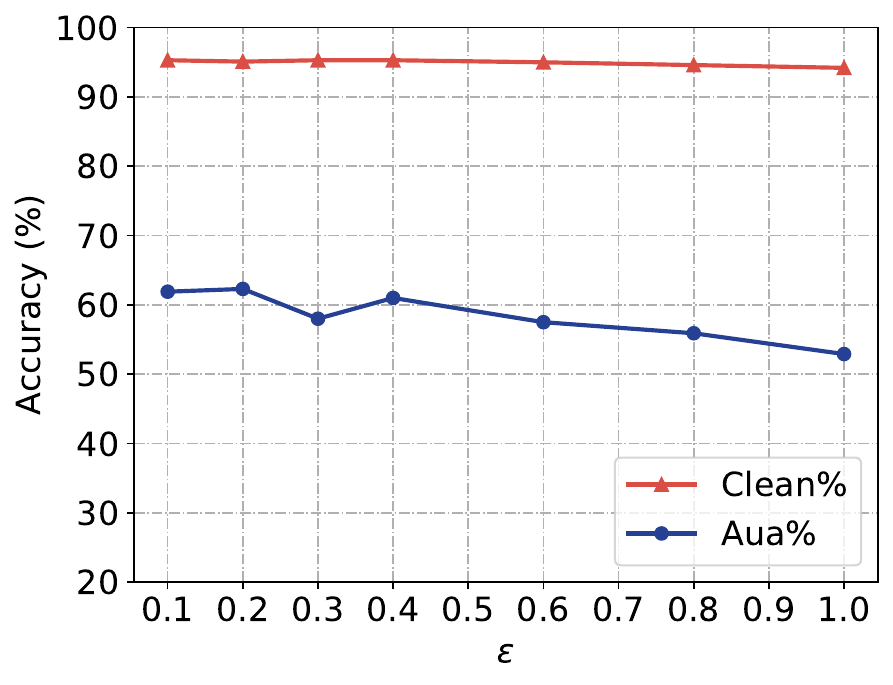}}
    \subfigure[\textit{QNLI}]{\label{fig:epsilon:qnli}\includegraphics[width=.32\textwidth]{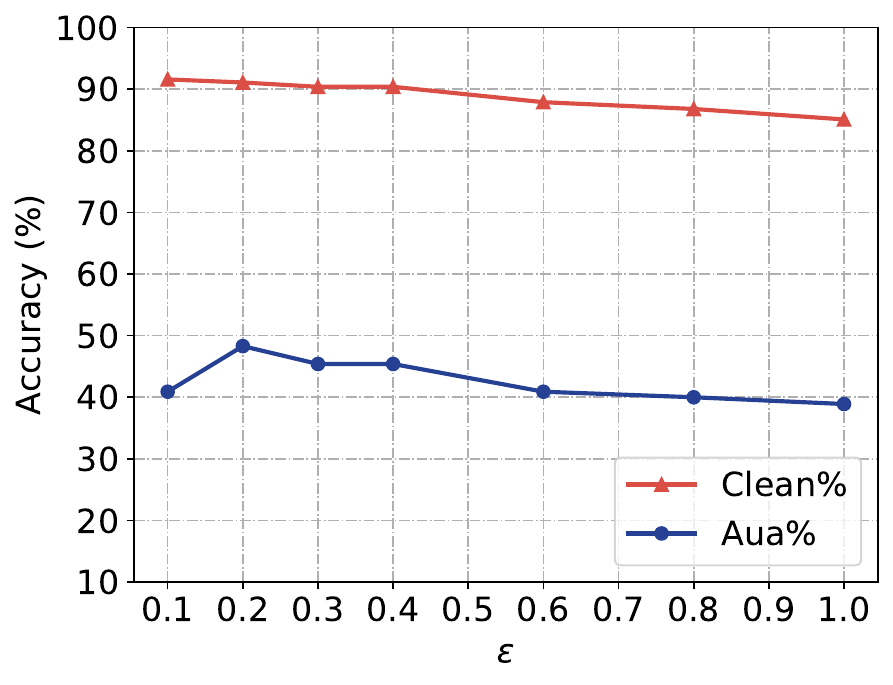}}
    \caption{The impact of hyper-parameter \(\epsilon\) on the performance of FAT across the three datasets.}
    \label{fig:epsilon}
\end{figure*}

\begin{figure*}[tb]
    \centering
    \subfigure[ \textit{IMDB}]{\label{fig:epoch:imdb}\includegraphics[width=.32\textwidth]{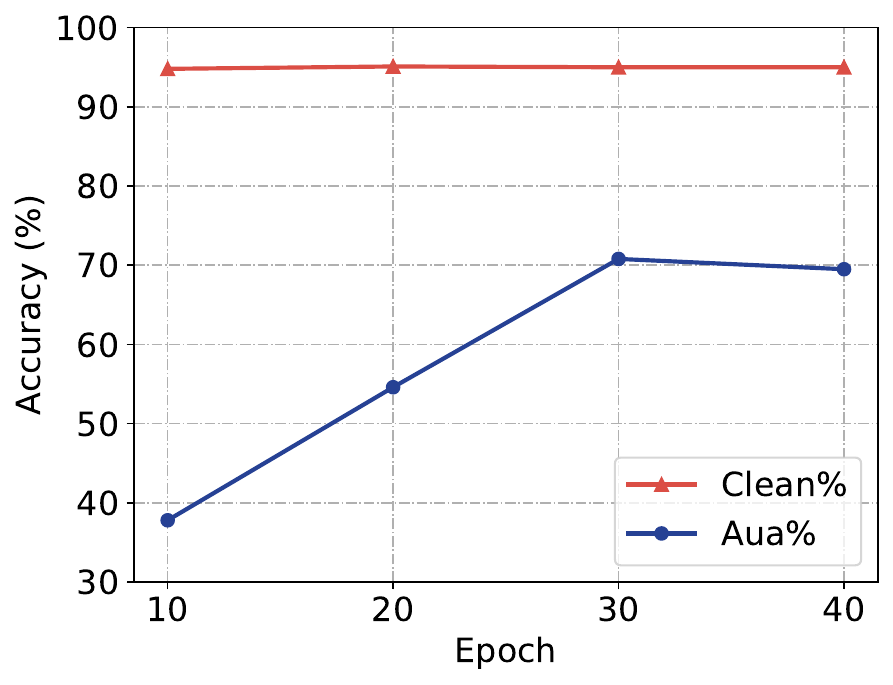}}
    \subfigure[ \textit{AGNEWS}]{\label{fig:epoch:agnews}\includegraphics[width=.32\textwidth]{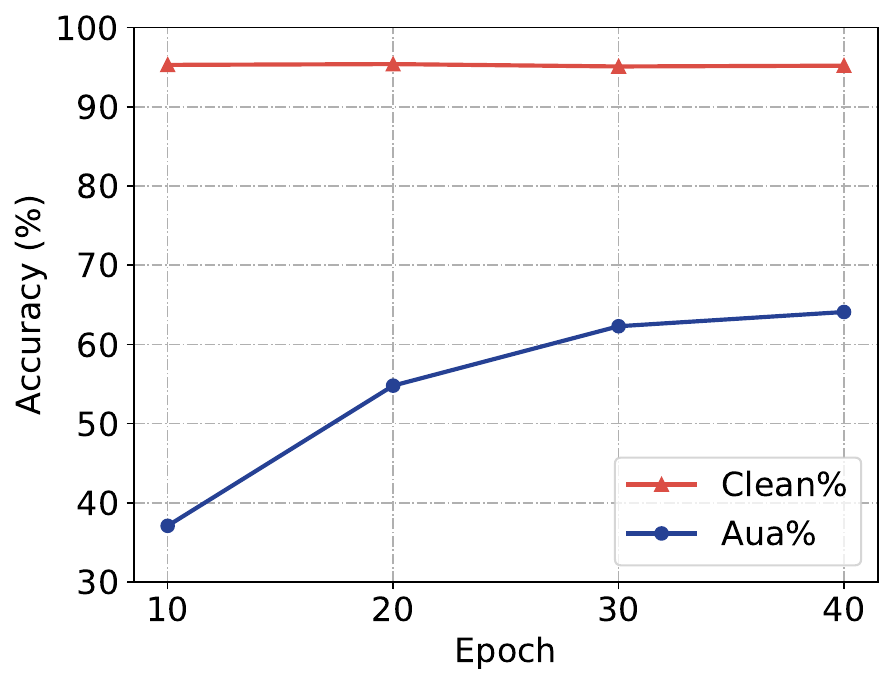}}
    \subfigure[\textit{QNLI}]{\label{fig:epoch:qnli}\includegraphics[width=.32\textwidth]{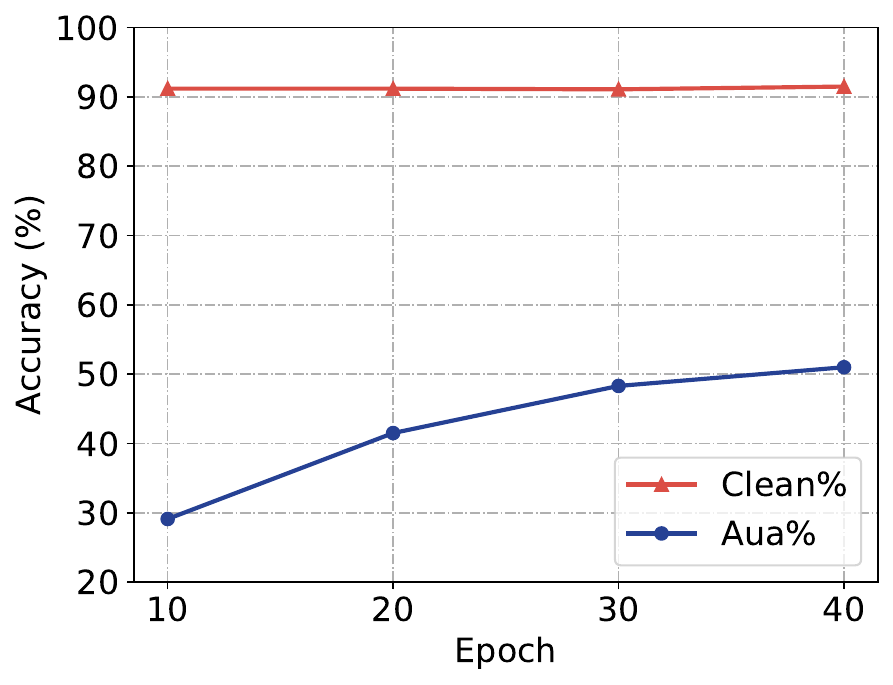}}
    \caption{The impact of training epochs on the performance of FAT across the three datasets.}
    \label{fig:epoch}
\end{figure*}

\subsection{Hyper-parameter Analysis}
\label{sec:hyper-parameter}

This subsection evaluates the impact of hyper-parameters on the performance of FAT. We focus on two metrics, the accuracy on the original test set, denoted by \textit{Clean\%}, and the robust accuracy under the TextFooler attack, denoted by \textit{Aua\%}. 

\subsubsection{Impact of Perturbation Magnitude} 

The hyper-parameter \(\epsilon\) in Eq.~(\ref{eq:fat}) controls the perturbation magnitude.  
To study the effect of \(\epsilon\) on FAT, we train the model with \(\epsilon=\) 0.1, 0.2, 0.3, 0.4, 0.6, 0.8, 1.0, respectively. 
The results are shown in Figure~\ref{fig:epsilon}. 
When \(\epsilon\) is set between 0.1 and 0.4, the clean accuracy remains almost constant on the three datasets. With the increase of \(\epsilon\), the clean accuracy decreases on the three datasets, especially on the \textit{QNLI} dataset. 
The robust accuracy begins to fluctuate within a small range, reaching a maximum when \(\epsilon=\) 0.2. As the value of \(\epsilon\) increases, the robust accuracy decreases significantly. To achieve the proper trade-off between clean accuracy and robustness, we set \(\epsilon=\) 0.2.
In addition, with the extensive range of \(\epsilon=\) 0.1 to 1.0, even the worst-case robust accuracy is significantly higher than all the defense baselines.



\subsubsection{Impact of Training Epoch}

We test the robustness of models trained with FAT for 10, 20, 30, and 40 epochs, respectively. The results are shown in Figure~\ref{fig:epoch}.
The accuracy of the original test set remains almost constant on the three datasets. At the beginning of training process, the model's robustness increases rapidly. After 30 training epochs, the rise in model's robustness slows significantly on the \textit{AGNEWS} and \textit{QNLI} datasets. On the \textit{IMDB} dataset, the model's robustness decreases slightly after 30 epochs, which we attribute to the fact that as the training progresses, the adversarial perturbations become too similar and overfitting occurs.
Thus, we train the model for 30 epochs to achieve a better trade-off between robustness and training time. 

\section{Conclusion}
Continuous AT method that directly adds perturbation to the embedding representation when training has the potential to provide robustness for NLP models under the synonym-unaware scenario. This work proposes a Fast Adversarial Training (FAT) method to boost adversarial robustness.
Specifically, through observing the characteristics of perturbation generation in the embedding space of NLP models, FAT adopts the single-step gradient ascent to generate adversarial perturbation and fully utilizes historical training information by initializing the perturbation along the perturbation direction of the previous epoch.
Extensive experiments on the BERT model using three datasets demonstrate that FAT outperforms the defense baselines under various adversarial attacks of different  perturbation granularity and model visibility by a clear margin. 
Since textual adversarial attacks have various settings of synonym candidates and perturbation budget, the proposed FAT, which does not introduce any human-prescribed linguistic rules or access the attackers' synonyms, is easy and practical.

Unlike the image domain, continuous AT has long been ignored as a powerful textual adversarial defense. Our experiments demonstrate that it is an efficacious and synonym-unaware defense method. On the one hand, we hope continuous AT can be considered as a strong baseline in 
future relevant research. 
On the other hand, since continuous AT bridges the gap between image and text adversarial training, some AT methods thoroughly studied in the image domain will be applied to the text domain in our future work.

\section{Acknowledgments}
This work is supported by National Natural Science Foundation of China (62076105, U22B2017) and International Cooperation Foundation of Hubei Province, China (2021EHB011).

\section*{Bibliographical References}\label{sec:reference}

\bibliographystyle{lrec-coling2024-natbib}
\bibliography{lrec-coling2024-example}

\end{document}